\documentclass{article}

\usepackage[preprint]{corl_2026} 
\usepackage{color}
\usepackage[usenames,dvipsnames,table]{xcolor}
\usepackage{times}
\usepackage{graphics} 
\usepackage{epsfig} 
\usepackage{mathptmx} 
\usepackage{times} 
\usepackage{amsmath} 
\usepackage{amssymb}  
\usepackage{multirow}
\usepackage{multicol}
\usepackage{enumerate}
\usepackage{tabularx}
\usepackage{hyperref}
\usepackage{pifont}
\usepackage{booktabs}
\usepackage{algorithm}
\usepackage{amsmath}
\usepackage{adjustbox}
\usepackage{wrapfig}
\usepackage{anyfontsize}
\usepackage[most]{tcolorbox}
\usepackage{lipsum} 
\hypersetup{
    colorlinks=true,
    linkcolor=MidnightBlue,
    filecolor=magenta,      
    urlcolor=MidnightBlue,
    citecolor=MidnightBlue,
} 
\usepackage[font=small,labelfont=bf]{caption}
\usepackage{xspace}
\newcommand{\sys}{AnyDexRT\xspace}

\title{\sys: Calibration-Efficient Dexterous Hand Retargeting with Few-Shot Human Guidance}

\author{
  Chenxi Wang$^{*,1}$ \quad Ying Feng$^{*,2,3}$ \quad Hongjie Fang$^{2,\dagger}$ \quad Shangning Xia$^1$ \\ \textbf{Lixin Yang}$^2$\quad \textbf{Chuan Wen}$^2$\quad \textbf{Cewu Lu}$^{1,2,3,\dagger}$ \\
  $^1$Noematrix\quad $^2$Shanghai Jiao Tong University\quad $^3$Shanghai Innovation Institute\\
  $^*$Equal Contribution\quad $^\dagger$Corresponding Authors
}

\begin{document}
\makeatletter
\let\@oldmaketitle\@maketitle
\renewcommand{\@maketitle}{\@oldmaketitle
\centering
\includegraphics[width=\linewidth]{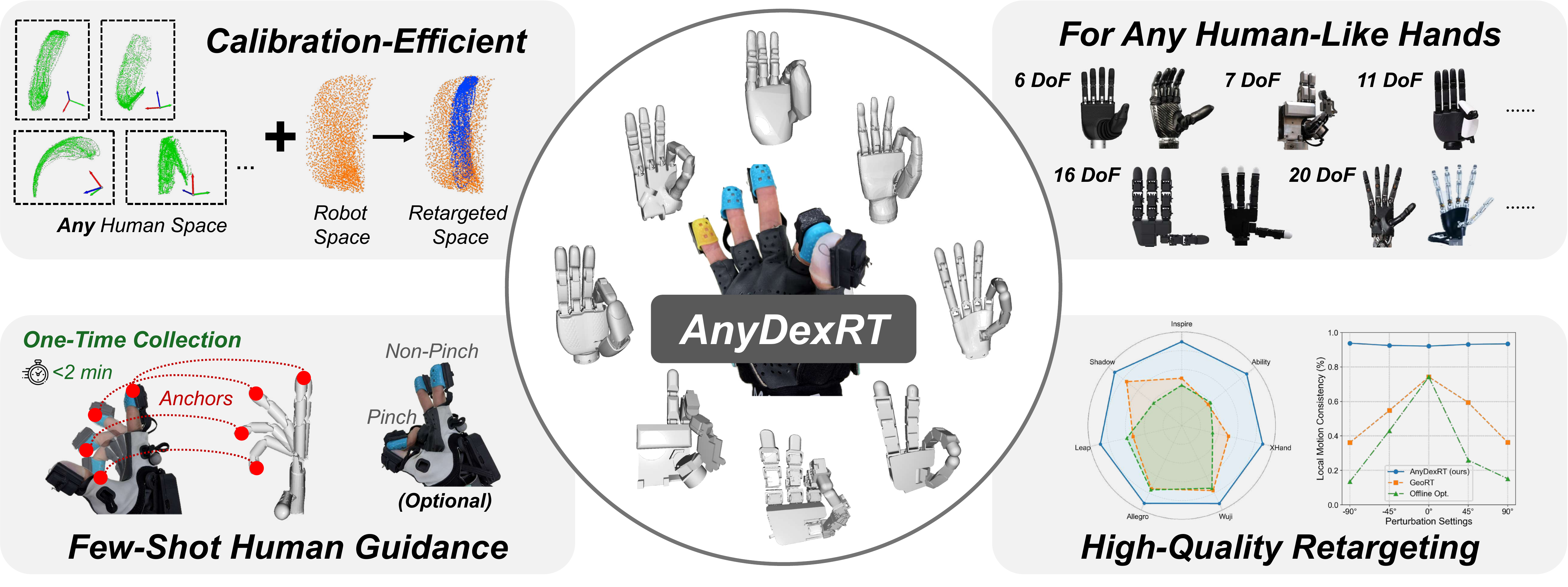}
\vspace{-0.4cm}
\captionof{figure}{\textbf{\sys System}. \sys is a calibration-efficient dexterous hand retargeting system. With few-shot human guidance, our system achieves high-quality retargeting across diverse human-like hands.}
\label{fig:teaser}
}%
\makeatother

\maketitle


\begin{abstract}
    Teleoperation is a key interface for controlling dexterous robotic hands and collecting demonstrations for imitation learning. Its effectiveness largely depends on kinematic retargeting, which maps operator hand motions to feasible and intuitive robot hand motions. Existing methods often require hand-crafted objectives, precise calibration, or global shape matching between human and robot hand spaces, making them sensitive to hand-specific tuning and less reliable across different dexterous hands. We propose \sys, a calibration-efficient retargeting method for intuitive dexterous teleoperation across human-like dexterous hands. 
    \sys combines self-supervised fingertip correspondence learning with few-shot human guidance to anchor the mapping in task-relevant regions, and further refines pinch-related poses using a contact classifier. 
    Experiments on diverse dexterous hands and real-world teleoperation tasks show that \sys improves retargeting quality, reduces manual tuning, and provides more intuitive and efficient control than prior retargeting methods.
    Project website: \url{https://chenxi-wang.github.io/projects/anydexrt}.
\end{abstract}

\keywords{Dexterous Hand Retargeting, Teleoperation, Dexterous Manipulation} 


\section{Introduction}\label{sec:intro}

Dexterous manipulation is a key capability for general-purpose robots, enabling rich and adaptive physical interactions with objects and environments~\cite{dex-survey, dexop, an2025dexterous}.
Compared with grippers, dexterous hands provide greater kinematic expressiveness and contact versatility, supporting diverse grasping~\cite{anydexgrasp, unidexgrasp}, in-hand object reorientation~\cite{chen2025vegetable, chen2023visual}, and contact-rich behaviors~\cite{li2025adaptive, suh2025dexterous} in unstructured environments. However, the same expressiveness that makes dexterous hands powerful also makes them difficult to control.
Their high-dimensional action spaces, joint couplings, and contact dynamics make it challenging to manually design effective hand motions.
Teleoperation provides a natural interface for accessing human dexterity, where an operator controls a robotic hand through their own hand motions while observing the robot's response in real time~\cite{anyteleop, robotic_telekinesis, geort}.
In this process, retargeting translates operator hand kinematics into feasible and intuitive robot hand motions.
The resulting teleoperated interactions can further serve as high-quality demonstrations for imitation learning~\cite{ohra, unidex, xlvla}, making retargeting important for both real-time control and data-driven robot learning.

As this interface, retargeting should not be reduced to direct pose matching between human and robot hands.
Because the operator and robot hands may differ in scale, motion range, joint coupling, and feasible configurations~\cite{retarget_review, xin2026analyzing}, a natural operator motion may not directly correspond to a natural robot motion.
A practical retargeting algorithm should therefore satisfy three requirements:
\begin{enumerate}
\item[\textbf{(R1)}] \textbf{Intuitiveness.} It should preserve the operator's motion intent and produce feasible, predictable robot motions that are easy to control.
\item[\textbf{(R2)}] \textbf{Calibration Efficiency.} It should reduce dependence on precise calibration or manually tuned hyperparameters such as scale factors, offsets, and task weights.
\item[\textbf{(R3)}] \textbf{Generality.} The same operator hand space should control different human-like dexterous hands without hand-specific redesign or handcrafted tuning.
\end{enumerate}

Existing methods still fall short of these requirements.
Traditional retargeting methods~\cite{anyteleop, dexpilot, wuji2026retargeting, dexmv, open-television, ace} often rely on inverse kinematics with hand-crafted task vectors or keypoints, requiring careful calibration of coordinate frames, vector origins, scale factors, and objective weights.
Neural variants~\cite{robotic_telekinesis, heng2026humdex, li2019vision} reduce runtime cost by imitating such retargeting pipelines, but remain tied to calibrated targets and hand-specific assumptions.
GeoRT~\cite{geort} instead learns correspondence by globally aligning the reachable fingertip spaces of human and robot hands.
However, global shape matching can be problematic when dexterous hands contain redundant feasible regions with no natural counterpart in the operator hand space, potentially distorting task-relevant mappings and yielding geometrically plausible but less intuitive teleoperation.

To this end, we propose \sys, a calibration-efficient retargeting method that enables intuitive dexterous teleoperation across different human-like dexterous hands with few-shot manual tuning.
\sys learns fingertip-level correspondence via self-supervised shape matching and uses few-shot human guidance to anchor the mapping in task-relevant regions, reducing ambiguity caused by redundant robot-hand motion spaces.
It further refines pinch-related poses with a contact classifier to improve the performance of pinch motions.
These designs reduce hand-specific calibration and tuning while maintaining an intuitive operator-to-robot mapping across different hands.
With a single, one-time data collection process, an operator can teleoperate multiple robotic hands without additional calibration.
Experiments across multiple dexterous hands and diverse real-world teleoperation tasks demonstrate that \sys improves retargeting quality, reduces manual tuning effort, and provides more intuitive teleoperation than prior methods.

\section{Related Works}

\subsection{Kinematic Retargeting for Dexterous Hands}
Kinematic retargeting converts human hand motion signals into robot joint commands for dexterous hands.
Some systems use hand-specific gloves~\cite{liu2017glove, liu2019high} or exoskeletons~\cite{dexop, dexumi} whose kinematics are designed to match the target hand.
While direct and responsive, such hardware is tied to a particular morphology and does not easily generalize to hands with different scales, proportions, or joint layouts.
Most systems instead decouple sensing from retargeting: they estimate human hand poses from vision-based trackers~\cite{dexmv, dexpilot, anyteleop, ace, open-television, robotic_telekinesis, ding2025bunny}, motion-capture gloves~\cite{heng2026humdex, dqrise, bidex, dexcap, bytedexter}, or other wearable sensors~\cite{doglove, chong2021learning}, and then pass fingertip poses to a retargeting module.

For the retargeting module, traditional optimization-based methods~\cite{antotsiou2018task, anyteleop, dexpilot, bytedexter, wuji2026retargeting, dexmv, open-television, ace, doglove, ding2025bunny, dexcap} manually define a set of task-vector pairs $\{(\mathbf{v}_i^H,\mathbf{v}_i^R)\}$ for the human and robotic hands, and solve for robot hand configurations by minimizing objectives such as $\sum_i\left\|\alpha_i\mathbf{v}_i^H - \mathbf{v}_i^R\right\|^2$, where $\{\alpha_i\}$ denotes the scaling factors for each finger. Several methods~\cite{chong2021learning, wuji2026retargeting} further introduce hand-specific objectives or constraints for particular robot hands, but such designs often lack generality. In practice, online optimization is commonly stabilized with joint limit, smoothness, or temporal regularization terms, while offline neural approaches~\cite{robotic_telekinesis, heng2026humdex, li2019vision} train networks to accelerate the retargeting process. However, when these networks are trained from calibrated optimization targets, they still inherit the assumptions and tuning requirements of the underlying retargeting formulation.

Recent work has explored learning human-robot correspondence with less manual correspondence design~\cite{geort,xin2026analyzing}. For example, GeoRT~\cite{geort} reduces manual correspondence design by treating retargeting as space alignment and amortizing the mapping into a neural policy.
However, its formulation introduces several practical limitations, including the use of a frozen neural forward-kinematics model, wrist-frame motion preservation that can be affected by fingertip misalignment, and bidirectional Chamfer matching over global reachable spaces.
These factors can make the learned mapping sensitive to training samples, calibration errors, and redundant robot-hand motion regions.

\subsection{Data Collection for Dexterous Manipulation}
High-quality demonstrations are essential for learning dexterous manipulation skills, but remain costly to collect for complex hand behaviors.
Prior work leverages specialized hardware~\cite{tilde, dexop, dexumi}, gloves~\cite{dexcap}, or in-the-wild human data~\cite{airexo, airexo2, umi}, yet such data may exhibit visual gaps from robot executions and kinematic gaps across different dexterous hands.
KineDex~\cite{kinedex} collects data directly on dexterous hands through kinesthetic teaching, but human-guided collection can introduce occlusions and becomes difficult when the target hand differs substantially from the human hand in scale.
Teleoperation therefore remains a primary paradigm for capturing expert behavior.
Some methods simplify control with predefined motion patterns, grasp primitives, or low-dimensional action spaces~\cite{hato, typetele}, trading expressiveness for lower operator burden.
Recent systems instead use direct hand-motion interfaces and kinematic retargeting to capture richer contact-rich behaviors~\cite{anyteleop, dexpilot, dexmv, open-television, ace, doglove, ding2025bunny, robotic_telekinesis, heng2026humdex}.
In these systems, retargeting directly affects demonstration quality: an unintuitive or poorly calibrated mapping can make teleoperation difficult and produce unnatural robot motions, limiting the usefulness of collected data for downstream imitation learning.

\section{AnyDexRT}

In this section, we present \sys, a calibration-efficient pipeline for mapping operator fingertip motions to robot hand commands.
After formulating the retargeting problem and assumptions (\S\ref{sec:preliminary}), we introduce how \sys learns fingertip correspondence through self-supervised shape matching (\S\ref{sec:self-supervised}), anchors task-relevant regions with few-shot human guidance (\S\ref{sec:anchor}), and refines pinch-related poses using a contact classifier (\S\ref{sec:pinch}). The overview of the system is shown in Fig.~\ref{fig:anydexrt}.

\subsection{Preliminary}\label{sec:preliminary}
For a hand with $F$ fingertips and $D$ joints, let $\mathbf{C}^H$,$\mathbf{C}^R$ $\subset \mathbb{R}^{F\times 3}$ be the set of 3D human and robot fingertip positions respectively, and $\mathbf{J}^R\subset \mathbb{R}^D$ be the set of the robot hand joints. We aim to find a mapping $f: \mathbf{C}^H \rightarrow \mathbf{J}^R$ from human fingertips to robot hand joints. The mapping $f$ can be modeled as a composite of two functions, \textit{i.e.}, $f = f_s \circ f_m$, where $f_m: \mathbf{C}^H\rightarrow \mathbf{C}^R$ maps human fingertip positions to robot fingertip positions, and $f_s: \mathbf{C}^R\rightarrow \mathbf{J}^R$ solves robot hand joints using robot fingertip positions.

To simplify the formulation while preserving generality, we make the following assumptions:

\begin{figure*}
    \centering
    \includegraphics[width=\linewidth]{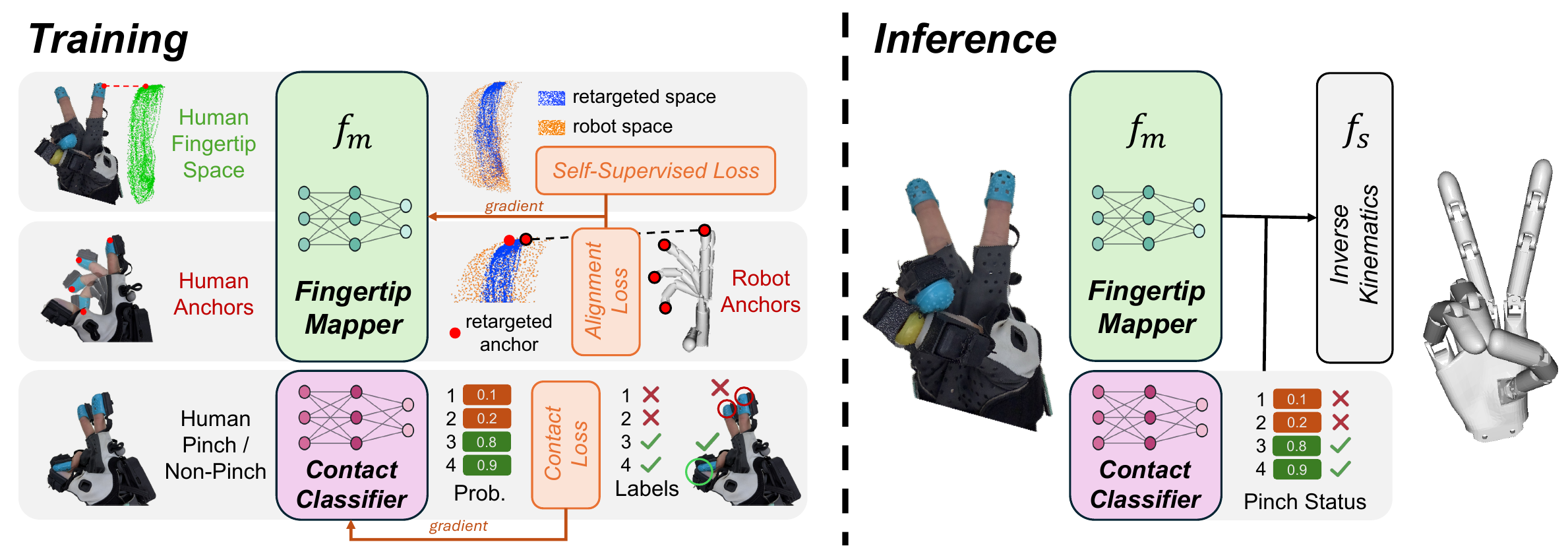}\vspace{-0.1cm}
    \caption{\textbf{Overview of \sys.} \textit{(Left)} During training, \sys learns a fingertip mapper from collected human fingertip samples with few paired human-robot anchors as guidance, and trains a contact classifier using collected or generated pinch/non-pinch poses. \textit{(Right)} During deployment, the fingertip mapper produces retargeted targets, which are refined by the contact classifier and converted into robot joint commands.}
    \label{fig:anydexrt}\vspace{-0.4cm}
\end{figure*}

\begin{enumerate}
    \item[\textbf{(A1)}] \textbf{The robotic hand is structurally similar to the human hand.} For human-like dexterous hands, stable finger coupling and low-dimensional postural synergies~\cite{rijpkema1991computer,santello1998postural,roel2015a} allow fingertip positions and reference joint angles to sufficiently constrain inverse kinematics (IK), so $f_s$ can be approximated as a one-to-one mapping.

    \item[\textbf{(A2)}] \textbf{The human fingertip motion space can be covered by the robotic fingertip space after a suitable geometric transformation.} This coverage is only required from human to robot, ensuring that natural human manipulation motions can be retained, while redundant robot-only regions need not be covered by the human hand.
\end{enumerate}

In this work, we mainly focus on the implementation of the fingertip mapper $f_m$. The function $f_s$ can be implemented through inverse kinematics, nearest neighbor search, or neural networks.

\subsection{Fingertip Mapping by Self-Supervised Shape Correspondence}\label{sec:self-supervised}
Fingertip trajectories reside on a manifold, which allows hand retargeting to be interpreted as shape correspondence between human and robotic fingertips. This correspondence can be learned in a self-supervised manner using several criteria.

\begin{figure*}
    \centering
    \includegraphics[width=\linewidth]{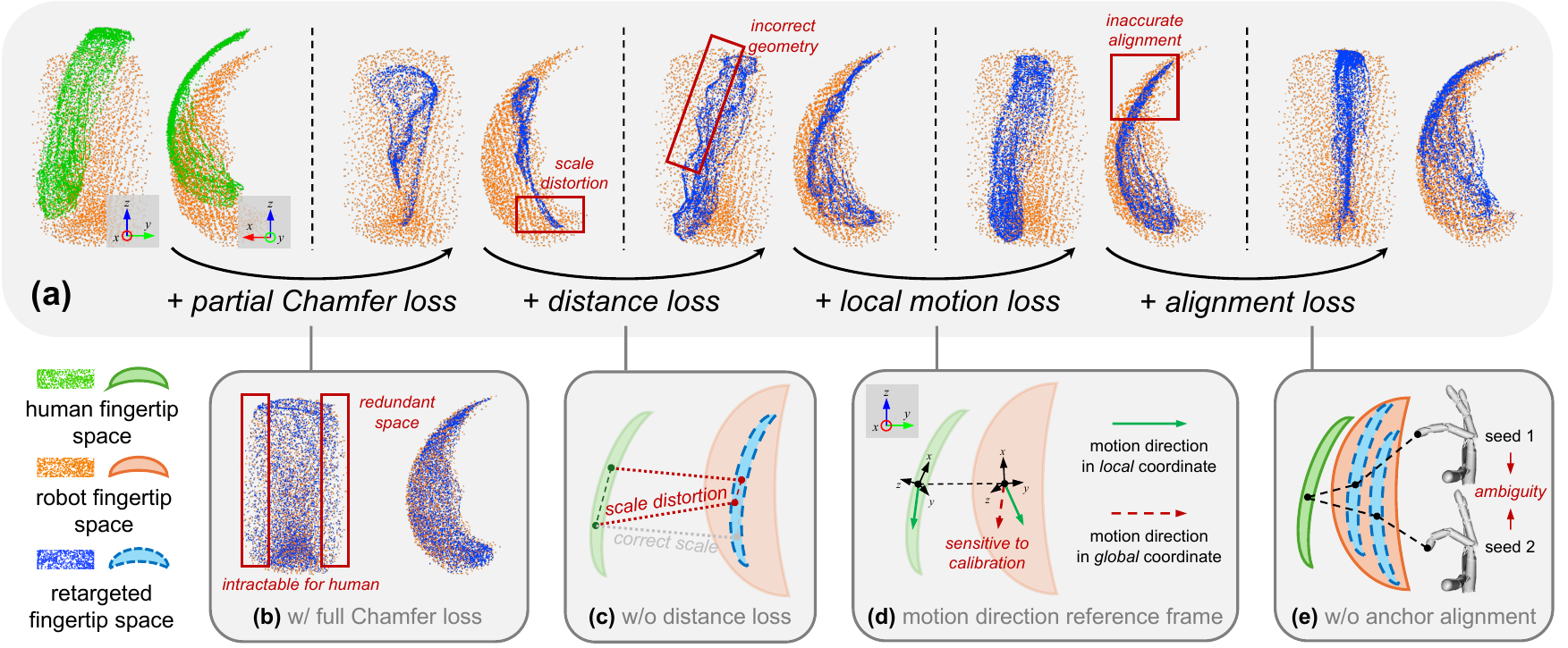}
    \caption{\textbf{Retargeting Objectives.}
\textbf{(a)} Combining all objectives produces a geometrically consistent and intuitive mapping from the human fingertip space to the robot fingertip space.
\textbf{(b)} Full Chamfer loss can force unnatural coverage of redundant robot fingertip regions.
\textbf{(c)} Distance loss preserves the geometric structure of the retargeted space and reduces mapping distortion.
\textbf{(d)} Local motion preservation encourages consistent motion directions between human and robot fingertips, and is less sensitive to calibration compared to global motion preservation.
\textbf{(e)} Few-shot anchor alignment resolves mapping ambiguity and stabilizes retargeting.}
    \label{fig:criterion}\vspace{-0.4cm}
\end{figure*}

\paragraph{Partial Correspondence} The Chamfer loss is commonly used for shape correspondence~\cite{chamfer, haoqiang2017a}, and has been adopted in prior retargeting work~\cite{geort}.
Its bidirectional formulation encourages full coverage between the mapped human fingertip space $f_m^i(\mathbf{C}^{H,i})$ and the robot fingertip space $\mathbf{C}^{R,i}$ for the $i$-th finger.
However, this assumption is not always suitable for dexterous hand retargeting.
According to \textbf{(A2)}, not every feasible robotic fingertip position has a natural counterpart in the human fingertip space.
For highly dexterous hands with redundant feasible regions, forcing $f_m^i(\mathbf{C}^{H,i})$ to cover the entire $\mathbf{C}^{R,i}$ may distort the learned mapping and produce an unnatural spatial distribution, as illustrated in Fig.~\ref{fig:criterion}(b). We therefore employ a partial Chamfer loss~\cite{xin2021cycle4completion}:
\begin{equation}
    \mathcal{L}_\text{P-Chamfer}(\mathbf{C}^{H,i}, \mathbf{C}^{R,i}) = \frac{1}{|\mathbf{C}^{H,i}|}\sum_{j=1}^{|\mathbf{C}^{H,i}|}\min_{k} \left\|f_m^i(x^{H,i}_j) - x^{R,i}_k\right\|,
\end{equation}
where $\left\|\cdot\right\|$ stands for L2 normalization. This asymmetric objective maps the human fingertip space into the feasible robot fingertip space without requiring the robot space to be fully covered, thereby preserving the structure of natural operator motions while reducing distortions caused by redundant robot-hand regions.

\paragraph{Distance Preservation} Although the partial Chamfer loss enforces a one-way mapping, it does not adequately preserve the original geometric distribution in $\mathbf{C}^{H}$. Fig.~\ref{fig:criterion}(c) illustrates an example of the geometric distortion. To mitigate this effect, we introduce a pairwise distance preservation objective. For any $x^{H,i}_{j_1},x^{H,i}_{j_2}\in\mathbf{C}^{H,i}$, we have
\begin{equation}
    \mathcal{L}_\text{dist}(\mathbf{C}^{H,i}) = \frac{1}{|\mathbf{C}^{H,i}|\left(|\mathbf{C}^{H,i}|-1\right)}\sum_{j_1\neq j_2}\left(\left\| f_m^i(x^{H,i}_{j_1}) - f_m^i(x^{H,i}_{j_2}) \right\| - \left\| x^{H,i}_{j_1} - x^{H,i}_{j_2} \right\|\right)^2.
\end{equation}
This objective regularizes the distribution of $f_m(\mathbf{C}^H)$ and alleviates geometric distortion.

\paragraph{Local Motion Preservation}
Partial correspondence aligns fingertip spaces, but does not guarantee that local motion directions are preserved.
GeoRT~\cite{geort} imposes directional consistency in the global hand-base frame, implicitly assuming well-calibrated fingertip measurements.
In practice, glove-sensor calibration errors may introduce translational or rotational discrepancies between measured and actual fingertip positions, causing globally measured directions to be misaligned.
We therefore enforce motion preservation in local coordinate frames, where the relative movement directions of the two spaces remain more consistent, as illustrated in Fig.~\ref{fig:criterion}(d). Let $\mathbf{T}(x)$ denote the local coordinate frame at position $x$.
For a small perturbation $\Delta x$ around $x_j^{H,i}$, the induced displacement in the robot space is
$\Delta f_m^i(x_j^{H,i}) = f_m^i(x_j^{H,i}+\Delta x)-f_m^i(x_j^{H,i})$.
We define the local motion loss as
\begin{equation}
    \mathcal{L}_\text{motion}(\mathbf{C}^{H,i}) = - \frac{1}{|\mathbf{C}^{H,i}|}\sum_{j=1}^{|\mathbf{C}^{H,i}|}\left<\textbf{T}^{-1}\left(x^{H,i}_j\right)\frac{\Delta x}{\left\|\Delta x\right\|}, \textbf{T}^{-1}\left(f_m^i(x^{H,i}_j)\right)\frac{\Delta f_m^i(x^{H,i}_j)}{\left\|\Delta f_m^i(x^{H,i}_j)\right\|}\right>,
\end{equation}
where $\langle\cdot,\cdot\rangle$ denotes the vector inner product.
Since $f_m^i(x_j^{H,i})$ does not provide a local rotation, we assign it the rotation of its nearest neighbor in $\mathbf{C}^{R,i}$.
Together with $\mathcal{L}_\text{P-Chamfer}$ and $\mathcal{L}_\text{dist}$, this local motion loss encourages retargeting behavior that is less sensitive to calibration errors and more consistent with operator intent.

\subsection{Geometric Alignment with Few-Shot Human Guidance}\label{sec:anchor}
Self-supervised learning allows the model to learn human-robot shape correspondence from unpaired fingertip samples, but the solution is not unique.
As illustrated in Fig.~\ref{fig:criterion}(e), the same human fingertip distribution $\mathbf{C}^H$ may be mapped to multiple plausible regions in the robot space $\mathbf{C}^R$, leading to unstable training and mappings that are sensitive to random sampling or initialization.
Such ambiguity can also introduce translation or scale offsets in fingertip positions, resulting in unnatural joint configurations and less reliable teleoperation.

To anchor the correspondence without requiring large-scale paired data, we introduce few-shot human guidance to the system.
Operators imitate a small set of reference gestures shown in Fig.~\ref{fig:anydexrt}, from which we collect paired human-robot fingertip anchors; the collection process is described in Appendix.
Given these anchors, we define the alignment loss as
\begin{equation}
    \mathcal{L}_\text{align}(\mathbf{C}^{H,i}, \mathbf{C}^{R,i}) = \frac{1}{M} \sum_{j=1}^M \left\| f_m^i(\bar{x}^{H,i}_j) - \bar{x}^{R,i}_j \right\|,
\end{equation}
where $\bar{x}^{H,i}_j$ and $\bar{x}^{R,i}_j$ denote the $j$-th paired anchor for the $i$-th finger, and $M$ is the number of anchors.
This sparse supervision resolves mapping ambiguity while preserving the scalability of self-supervised correspondence learning.
It also enables personalized retargeting, as operators can specify preferred scales and reference poses with only a few intuitive gestures.

\subsection{Pinch Pose Refinement using Contact Classifier}\label{sec:pinch}
\begin{wrapfigure}{r}{0.3\linewidth}\vspace{-0.9cm}
    \centering
    \includegraphics[width=\linewidth]{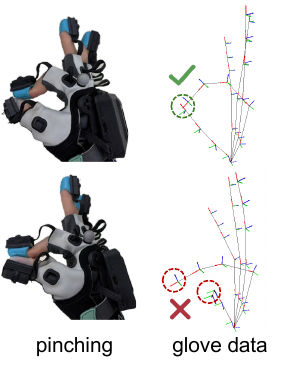}\vspace{-0.3cm}
    \caption{\textbf{Sensor Failure during Pinches.} The measured data fail to capture hand pinches.}
\label{fig:pinch}
\end{wrapfigure}
Grasping tiny objects requires higher retargeting accuracy, as even small positional errors can lead to task failure. This is especially critical for pinch motions, where precise fingertip contact is required~\cite{xin2026analyzing}. As shown in Fig.~\ref{fig:pinch}, due to sensor limitations and potential electromagnetic interference, the measured hand pose may still deviate significantly from the actual hand gesture, even when the operator makes clear fingertip contact. As a result, the measured pose may fail to capture the geometric characteristics of a pinch.

To address this issue, we no longer rely solely on fingertip position mapping to infer fine contact patterns. Since finger contact signals are easy to collect and annotate, and can directly reflect key manipulation intentions such as pinching, we instead train a classifier $f_c$ to recognize such contact patterns. Let $0$ be the thumb index and $i>0$ be the index of other fingers, we have
\begin{equation}
    \mathcal{L}_\text{contact}(\mathbf{C}^{H,i}) = \frac{1}{N}\sum_{j=1}^{N}\text{BCE}\left(y^i_j, f_c^i(x^{H,0}_{j},x^{H,i}_{j})\right),
\end{equation}
where $N$ denotes the number of training samples, $\text{BCE}(\cdot,\cdot)$ denotes the binary cross entropy loss, and $y^i_j\in\{0,1\}$ denotes the label of the $j$-th sample.

The contact classifier provides a more reliable retargeting signal for grasping tiny objects. During inference, when a human fingertip contact is detected, we search the neighborhood of the corresponding mapped robotic position for a robotic pinch pose, ensuring stable grasping.

\section{Experiments}

In this section, we assess \sys through simulation experiments and a real-world teleoperation evaluation.
The evaluation aims to verify whether \sys satisfies the three requirements introduced in \S\ref{sec:intro}, including \textbf{(R1)} intuitiveness, \textbf{(R2)} calibration efficiency, and \textbf{(R3)} generality, and to compare its retargeting quality, efficiency, and operator-perceived intuitiveness with prior methods.

\subsection{Simulation Experiments}

\paragraph{Setup.}
We conduct simulated experiments on seven human-like dexterous hands listed in Tab.~\ref{tab:results} to evaluate \textbf{(R3)}.
These hands span diverse kinematic structures and range from 6 to 20 DoFs, testing whether each method can adapt to different hand embodiments.
The same human data are used to train the models for all robotic hands.
To evaluate retargeting quality, we measure motion consistency~\cite{geort}, which reflects whether the robot hand responds consistently with the operator's motion intent \textbf{(R1)}.
We report global motion consistency (GMC) and local motion consistency (LMC).
GMC follows GeoRT~\cite{geort} and compares displacement directions in a shared coordinate frame, assuming an ideal, well-calibrated setting.
LMC compares directions in local frames, making it less calibration-dependent and more aligned with operator control.
We omit whole-space coverage as a main metric, since covering redundant robot regions does not necessarily improve teleoperation intuitiveness.
We compare \sys with a representative optimization-based retargeting method~\cite{anyteleop} (offline version) and the neural retargeting method GeoRT~\cite{geort}.
For methods involving stochastic training or sampling, we report results over 5 random seeds to evaluate their stability and sensitivity to initialization.

\begin{table}[t]
    \centering\footnotesize
    \setlength\tabcolsep{4pt}
    \begin{tabular}{c ll ll ll ll}
    \toprule
        \multirow{2}{*}{\begin{tabular}{c}\textbf{Retargeting}\\ \textbf{Method}\end{tabular}} & \multicolumn{2}{c}{\textbf{Inspire Hand}~\cite{inspire_hand}} & \multicolumn{2}{c}{\textbf{Ability Hand}~\cite{ability_hand}} & \multicolumn{2}{c}{\textbf{XHand}~\cite{xhand}} &  \multicolumn{2}{c}{\textbf{Wuji Hand}~\cite{wuji_hand}}\\ \cmidrule(lr){2-3} \cmidrule(lr){4-5} \cmidrule(lr){6-7} \cmidrule(lr){8-9}
        & \multicolumn{1}{c}{GMC} & \multicolumn{1}{c}{LMC} & \multicolumn{1}{c}{GMC} & \multicolumn{1}{c}{LMC} & \multicolumn{1}{c}{GMC} & \multicolumn{1}{c}{LMC} & \multicolumn{1}{c}{GMC} & \multicolumn{1}{c}{LMC} \\ \midrule
        Optimization~\cite{anyteleop}* & $53.9$ & $43.2$ & $47.7$ & $39.5$ & $46.9$ & $33.8$ & $80.2$ & $74.2$ \\
        GeoRT~\cite{geort} & $71.9_{\pm3.5}$ & $50.6_{\pm 3.6}$ & $65.0_{\pm 4.7}$ & $36.2_{\pm 11.2}$ & $75.4_{\pm 4.1}$ & $51.3_{\pm4.2}$ & $87.3_{\pm 2.6}$ & $77.0_{\pm4.6}$\\
        AnyDexRT \textit{(ours)}  & $\textbf{84.7}_{\pm 0.1}$ & $\textbf{92.1}_{\pm 0.1}$ & $\textbf{85.0}_{\pm 0.2}$ & $\textbf{90.5}_{\pm 0.1}$ & $\textbf{89.7}_{\pm 0.2}$ & $\textbf{94.8}_{\pm0.1}$ &  $\textbf{93.2}_{\pm 0.2}$ & $\textbf{94.1}_{\pm0.3}$\\
        \midrule
        \multirow{2}{*}{\begin{tabular}{c}\textbf{Retargeting}\\ \textbf{Method}\end{tabular}} & \multicolumn{2}{c}{\textbf{Allegro Hand}~\cite{allegro_hand}} & \multicolumn{2}{c}{\textbf{Leap Hand}~\cite{leap_hand}} & \multicolumn{2}{c}{\textbf{Shadow Hand}~\cite{shadow_hand}} &  \multicolumn{2}{c}{\cellcolor[HTML]{f2f2f2}\textbf{Average} (7 Hands)}\\ \cmidrule(lr){2-3} \cmidrule(lr){4-5} \cmidrule(lr){6-7} \cmidrule(lr){8-9}
        & \multicolumn{1}{c}{GMC} & \multicolumn{1}{c}{LMC} & \multicolumn{1}{c}{GMC} & \multicolumn{1}{c}{LMC} & \multicolumn{1}{c}{GMC} & \multicolumn{1}{c}{LMC} & \multicolumn{1}{c}{\cellcolor[HTML]{f2f2f2}GMC} & \multicolumn{1}{c}{\cellcolor[HTML]{f2f2f2}LMC} \\ \midrule
        Optimization~\cite{anyteleop}* & $89.5$ & $76.1$ & $70.8$ & $60.0$ & $45.0$ & $38.5$ & \multicolumn{1}{c}{\cellcolor[HTML]{f2f2f2} 62.0} & \multicolumn{1}{c}{\cellcolor[HTML]{f2f2f2} 52.2} \\
        GeoRT~\cite{geort} & $85.5_{\pm1.9}$ & $74.8_{\pm 1.6}$ & $73.4_{\pm 3.2}$ & $53.2_{\pm 8.8}$ & $89.8_{\pm 1.6}$ & $75.2_{\pm4.1}$ & \multicolumn{1}{c}{\cellcolor[HTML]{f2f2f2} 78.3} & \multicolumn{1}{c}{\cellcolor[HTML]{f2f2f2} 59.8} \\
        AnyDexRT \textit{(ours)}  & $\textbf{92.2}_{\pm 0.1}$ & $\textbf{94.0}_{\pm 0.1}$ & $\textbf{92.3}_{\pm 0.1}$ & $\textbf{93.2}_{\pm0.3}$ & $\textbf{90.8}_{\pm 0.2}$ & $\textbf{94.1}_{\pm 0.1}$ & \multicolumn{1}{c}{\cellcolor[HTML]{f2f2f2} \textbf{89.7}} & \multicolumn{1}{c}{\cellcolor[HTML]{f2f2f2} \textbf{93.3}} \\
        \bottomrule
    \end{tabular}\vspace{0.1cm}
    \caption{\textbf{Simulation Results ($\times 10^{-2}$) on Retargeting Quality.}  \sys achieves better global and local motion consistency across diverse dexterous hands. 
   *Deterministic methods do not have standard deviations.}\vspace{-0.4cm}
    \label{tab:results}\vspace{-0.4cm}
\end{table}

\paragraph{Retargeting Quality.} Tab.~\ref{tab:results} reports retargeting quality across seven dexterous hands.
\sys achieves strong performance across different hand embodiments, improving the average local motion consistency from $59.8\%$ to $90.2\%$.
This shows that \sys better preserves the operator's motion intent, supporting \textbf{(R1)}.
Although optimized for local consistency, \sys also maintains competitive global motion consistency under an ideal, well-calibrated shared frame.
Moreover, \sys achieves consistently high scores with small standard deviations across hands, while GeoRT exhibits larger variations across hands and random seeds.
These results show that \sys provides more stable and general retargeting, supporting \textbf{(R3)}.
We additionally evaluate the contact classifier on a separately collected testing set. It achieves 95.5\% accuracy, 98.7\% precision, and 92.3\% recall, demonstrating the reliability of the classifier in recognizing human pinch motions.

\begin{table}[t]
\begin{minipage}{0.3\linewidth}
    \centering\footnotesize
    \setlength\tabcolsep{2pt}
    \begin{tabular}{ccc}
        \toprule
        \textbf{Method} & \textbf{\# HP} & \textbf{Speed} (Hz) \\ \midrule
        \cite{anyteleop} & $\ge 10$ & 93.4\\
        \cite{geort} & 4 & 281.7\\
        \textit{ours} & \textbf{3} & \textbf{293.0}\\ \bottomrule
    \end{tabular}\vspace{0.1cm}
    \captionof{table}{\textbf{Retargeting Comparisons.} HP is hyperparameter.}\label{tab:comp}

    \begin{tabular}{ccc}
        \toprule
        & \textbf{Objective} & \textbf{LMC} ($\times 10^{-2}$) \\ \midrule
        & $\mathcal{L}_\text{P-Chamfer}$ & 4.1\\
        + & $\mathcal{L}_\text{dist}$ & 83.8\\
        + & $\mathcal{L}_\text{motion}$ & 89.1\\
        + & $\mathcal{L}_\text{align}$ & 94.1\\ \bottomrule
    \end{tabular}\vspace{0.1cm}
    \captionof{table}{\textbf{Ablation Results.}}\label{tab:ablation_criteria}\vspace{-0.4cm}
\end{minipage}
\hfill
\begin{minipage}{0.34\linewidth}
    \centering
    \includegraphics[width=0.97\linewidth]{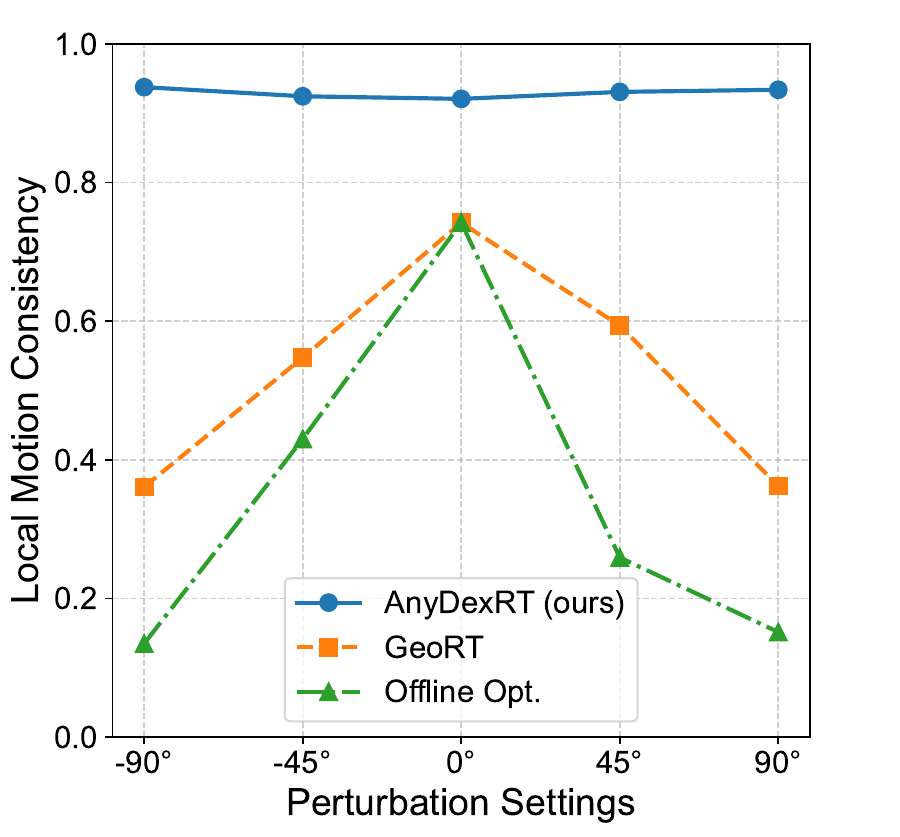}
    \captionof{figure}{\textbf{Calibration Sensitivity.} \sys maintains consistent retargeting quality, demonstrating its calibration-efficient capability.}
    \label{fig:calib}
\end{minipage}
\hfill
\begin{minipage}{0.34\linewidth}
    \centering
    \includegraphics[width=0.98\linewidth]{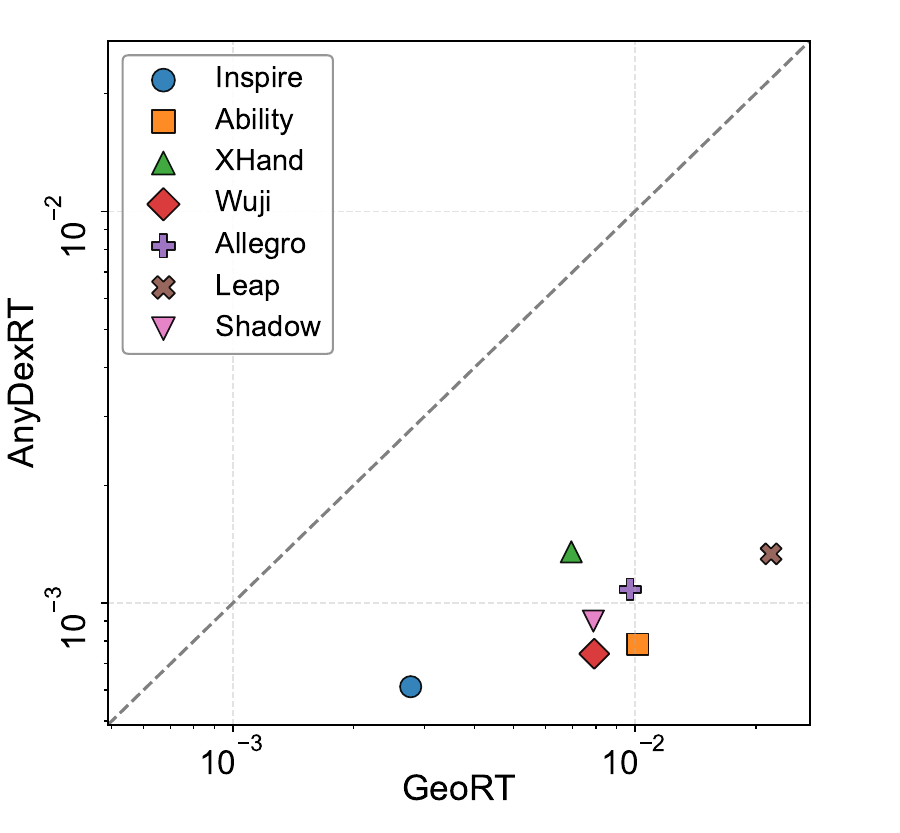}
    \captionof{figure}{\textbf{Retargeting Standard Deviations.} \sys shows strong mapping stability over GeoRT across different training initializations.}\label{fig:std}
\end{minipage}

\vspace{-0.8cm}
\end{table}
\paragraph{Manual Tuning Effort and Retargeting Efficiency.}
We further compare manual tuning effort and the runtime efficiency of different retargeting methods in Tab.~\ref{tab:comp}. Similar to GeoRT, \sys requires only 3 hyperparameters and runs at about 300Hz, outperforming the optimization baseline in both tuning effort and speed.
In practice, the default hyperparameters of \sys rarely need adjustment and can be directly reused across different hands.
These results show that \sys reduces tuning effort while maintaining real-time retargeting efficiency, supporting \textbf{(R2)}.

\paragraph{Calibration Sensitivity.}
We use Wuji Hand to evaluate calibration sensitivity.
To simulate frame misalignment between glove measurements and the robot hand, we rotate the input human data around the $y$-axis by $-90^\circ$, $-45^\circ$, $45^\circ$, and $90^\circ$.
We then calculate LMC under each perturbation to assess whether the mapping preserves operator motion intent despite calibration errors.
As shown in Fig.~\ref{fig:calib}, the optimization-based baseline and GeoRT degrade substantially under these perturbations, indicating their dependence on accurate coordinate alignment.
In contrast, \sys maintains stable retargeting quality across all rotations, demonstrating its calibration-efficient ability to adapt to misaligned hand coordinate systems and better satisfy \textbf{(R2)}.

\paragraph{Training Stability.} Fig.~\ref{fig:std} reports the training stability of \sys and GeoRT~\cite{geort}, which is computed by averaging the standard deviations of the retargeted positions over 5 random seeds. We can observe that \sys achieves orders-of-magnitude better stability than GeoRT under different training initializations across all hands. Such stable mapping ensures the reproducibility of the results and substantially reduces the difficulty of teleoperation.

\paragraph{Ablation.} We provide qualitative ablations in Fig.~\ref{fig:criterion}(a) and quantitative results on Wuji Hand in Tab.~\ref{tab:ablation_criteria}.
$\mathcal{L}_\text{dist}$ and $\mathcal{L}_\text{motion}$ improve LMC by preserving the global geometric structure and local motion directions, respectively.
Adding $\mathcal{L}_\text{align}$ further anchors the mapping with few-shot guidance, contributing to more natural and intuitive retargeting.

\subsection{Real-World Teleoperation Performance}

\begin{table}[t]
    \begin{minipage}{0.25\linewidth}
    \centering
    \includegraphics[width=\linewidth]{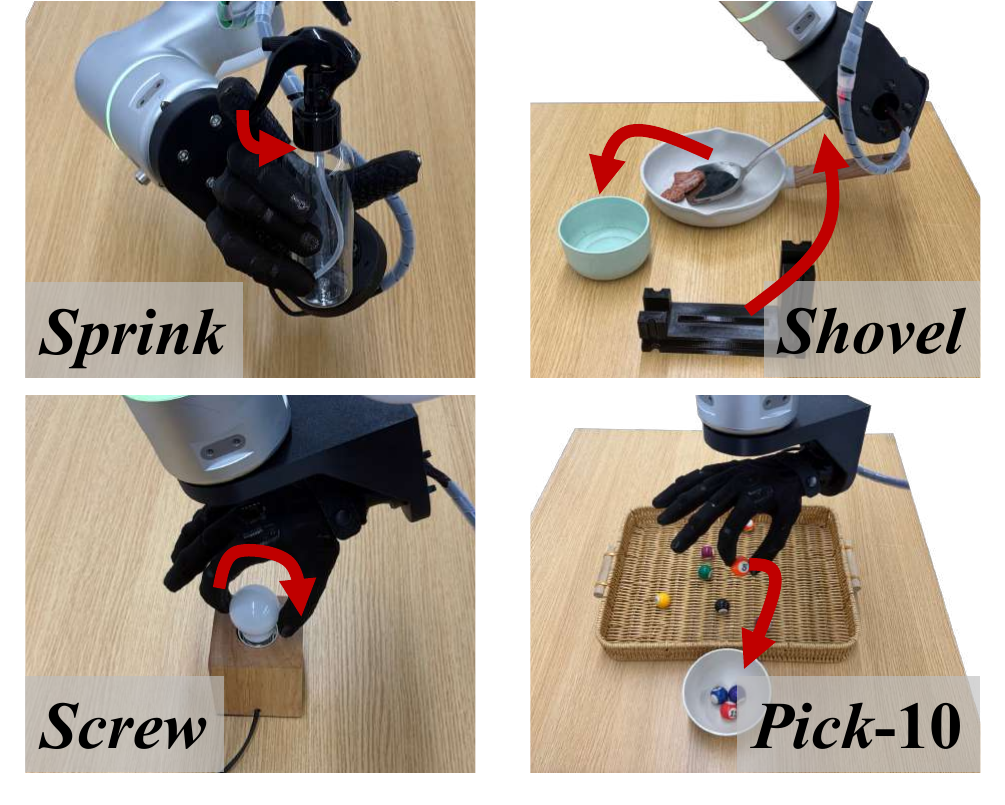}
    \end{minipage}
    \hfill
    \begin{minipage}{0.7\linewidth}
    \centering\footnotesize
    \setlength\tabcolsep{4pt}
    \begin{tabular}{c rrrr c}
    \toprule
    \multirow{2}{*}{\begin{tabular}{c}\textbf{Retargeting}\\ \textbf{Method}\end{tabular}} & \multicolumn{4}{c}{\textbf{Time per Episode} (s) $\downarrow$} & \multirow{2}{*}{\begin{tabular}{c}\textbf{Pinch} \\ \textbf{Success Rate} $\uparrow$\end{tabular}} \\ \cmidrule(lr){2-5}
    & \textbf{\textit{Sprink}} & \textbf{\textit{Screw}} & \textbf{\textit{Shovel}} & \textbf{\textit{Pick}-10} \\ \midrule
    Optimization~\cite{anyteleop} & 29.0 & 25.3 & 36.4 & 150.4 & 39.6\%\\
    GeoRT~\cite{geort} & 32.1 & 22.8 & 38.5 & 220.4 & 29.2\% \\
    \sys~\textit{(ours)} & \textbf{10.6} & \textbf{17.0} & \textbf{28.0} & \textbf{105.8} & \textbf{62.0}\%\\
        \bottomrule
    \end{tabular}
    \end{minipage}
\caption{\textbf{Real-World Teleoperation Performance.}
\textit{(Left)} Four real-world evaluation tasks, including dexterous tasks for assessing retargeting quality and tiny-object grasping tasks for assessing pinch control.
\textit{(Right)} \sys improves task efficiency and pinch success rates over prior retargeting methods, demonstrating its effectiveness in real-world teleoperation and its applicability to collecting dexterous manipulation data.}
    \label{tab:realworld}\vspace{-0.7cm}
\end{table}

\paragraph{Setup.}
We use a Flexiv Rizon 4 arm~\cite{flexiv_arm} equipped with a Wuji Hand~\cite{wuji_hand} as the robot platform. The operator wears a Manus glove~\cite{manus} and an HTC Vive Tracker~\cite{vive_tracker}. The glove provides fingertip poses for hand retargeting, while the tracker controls the robot arm. Each retargeting method maps the glove fingertip poses to real-time joint commands for the Wuji Hand. We evaluate four real-world tasks: spray-bottle triggering (\textbf{\textit{Sprink}}), light-bulb screwing (\textbf{\textit{Screw}}), steak shoveling (\textbf{\textit{Shovel}}), and small-ball picking (\textbf{\textit{Pick}-10}). These tasks span finger-specific actuation, grasping, tool-use, contact-rich behavior, and repetitive pinch manipulation.

\paragraph{Protocols.}
8 operators uninvolved in the project participated in the evaluation, including 3 experts and 5 non-experts. Each operator evaluated all methods, with method identities \textit{concealed} and the method order \textit{randomized} across operators. Before evaluation, each method was given an \textit{identical 10-min practice period}, after which the tasks were performed in a fixed easy-to-hard order. Each task consisted of 3 trials, with a per-trial timeout of 5 min for \textbf{\textit{Pick-10}} and 2 min for all other tasks.

\paragraph{Intuitiveness and Efficiency.}
As shown in Tab.~\ref{tab:realworld}, \sys achieves the shortest completion time on all tasks, suggesting more efficient and predictable teleoperation than the baselines.
The improvements are especially clear in \textbf{\textit{Sprink}} and \textbf{\textit{Shovel}}, which require accurate finger response and stable grasp adjustment.
These results show that \sys improves real-world dexterous control quality during teleoperation, supporting \textbf{(R1)}. More analyses are provided in the Appendix.

\paragraph{Pinch Performance.}
We evaluate pinch control using the \textbf{\textit{Pick}-10} task, which requires reliable thumb-finger contact to pick up ten small balls. The task is challenging because the balls are \textit{finger-sized} and have \textit{smooth, curved surfaces}, while 5 of the 8 users have no prior teleoperation experience. \sys achieves the highest pinch success rate, significantly outperforming both the optimization-based method and GeoRT. Replacing the balls with same-sized blocks increases the success rate by 30.0 percentage points (pp) to 92.0\%, further highlighting the difficulty of the original task. An ablation study on \textbf{\textit{Pick}-10} shows a 12.2 pp drop in success rate when the contact classifier is removed. These results demonstrate that contact-classifier refinement improves the reliability of contact-critical pinch motions and enables higher-quality dexterous data collection.

\begin{figure}
    \centering
    \includegraphics[width=\linewidth]{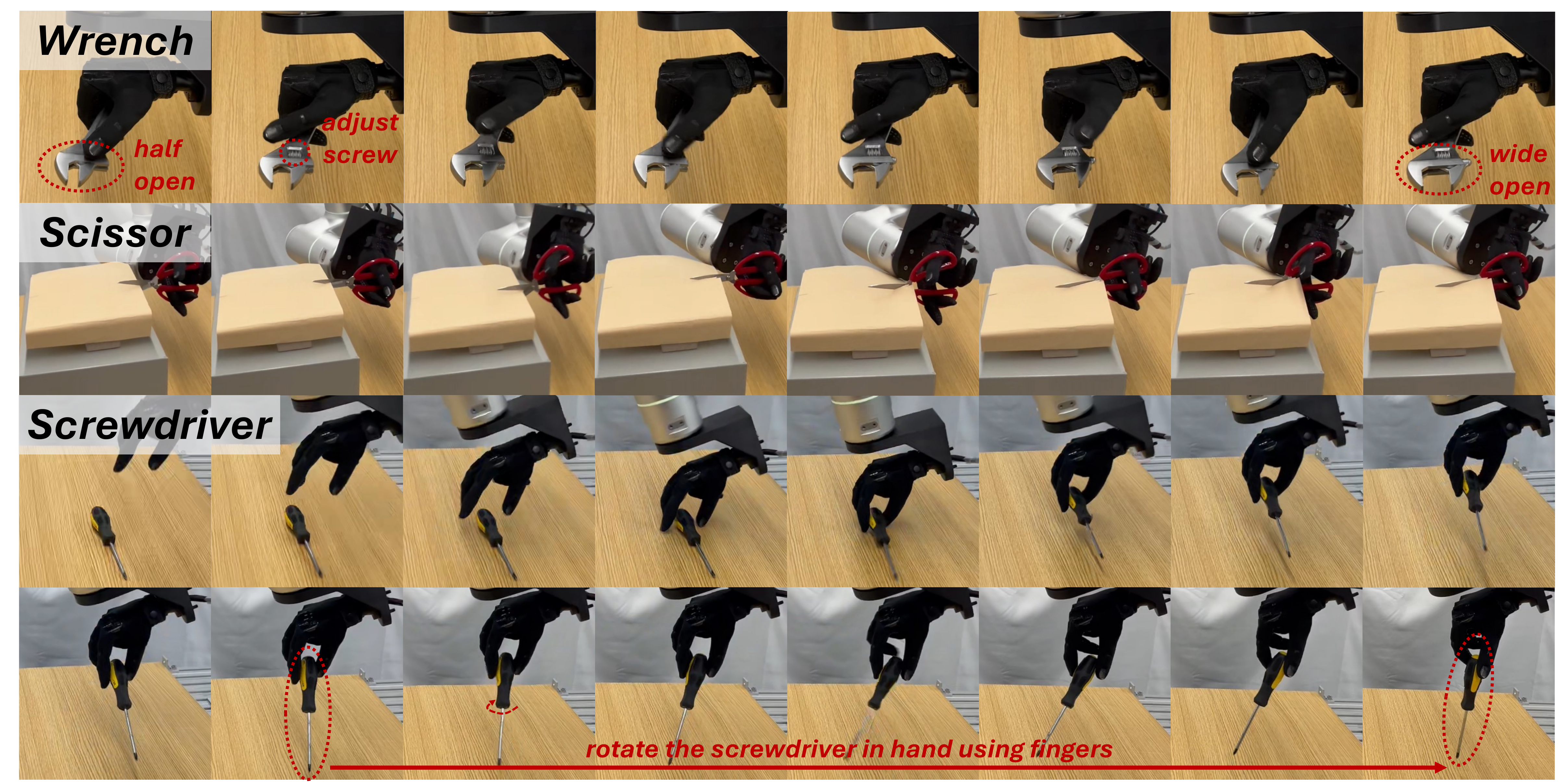}
    \caption{\textbf{Qualitative Real-World Teleoperation Performance on Additional Challenging Tasks.} A proficient operator uses \sys to teleoperate a Wuji hand for \textbf{\textit{Wrench}}, \textbf{\textit{Scissor}}, and \textbf{\textit{Screwdriver}} tasks.}\vspace{-0.2cm}
    \label{fig:additional-teleop}\vspace{-0.4cm}
\end{figure}

\paragraph{Qualitative Teleoperation Results on Additional Challenging Tasks} Fig.~\ref{fig:additional-teleop} presents qualitative results of utilizing \sys to complete the additional challenging teleoperation tasks, including opening a wrench (\textbf{\textit{Wrench}}), cutting a sponge with scissors (\textbf{\textit{Scissor}}), and grasping, lifting, and rotating a screwdriver (\textbf{\textit{Screwdriver}}). These results demonstrate the effectiveness of \sys in performing challenging, contact-rich tool-use tasks, even in the absence of force feedback.

\section{Conclusion and Limitations}
We present \sys, a calibration-efficient retargeting method for intuitive teleoperation across different human-like dexterous hands.
\sys learns fingertip mapping through self-supervised shape matching, while using few-shot human guidance to anchor task-relevant regions and reduce ambiguity caused by redundant robot fingertip spaces.
It further refines pinch-related poses with a contact classifier, improving the responsiveness and controllability of contact-critical motions.
Experiments across multiple dexterous hands demonstrate that \sys achieves strong retargeting quality with lower tuning effort and better stability than representative baselines.
Real-world teleoperation evaluations further show that \sys enables more efficient task completion and more reliable pinch control, supporting its applicability to dexterous teleoperation and data collection.

\paragraph{Limitations and Future Work.}
Although \sys reduces calibration and tuning effort, it still requires human-guided anchors; future work could adapt the mapping online from operator feedback.
Our contact-aware refinement currently focuses on pinch-related poses, while broader contact-rich behaviors may require richer contact models.
Finally, our real-world evaluation mainly validates teleoperation performance.
Training downstream manipulation policies with the collected data would further assess the value of \sys for dexterous data collection.

\section*{Acknowledgement}
This work was supported by the Fundamental and Interdisciplinary Disciplines Breakthrough Plan of the Ministry of Education of China, Jiangsu Science and Technology Major Project (BG2025035), Shanghai Artificial Intelligence Laboratory, XPLORER PRIZE grants.

We would like to thank Hao-Shu Fang from MIT for insightful discussions. We are also grateful to Shirun Tang, Junchao Zhang, Zelin Ye from Noematrix, and Zihao He from SJTU for their support during the real-world teleoperation evaluation, and to Shixuan Huang from Noematrix for designing the connector between the arm and the dexterous hand.


{\small
\bibliography{references}
}

\clearpage
\appendix
{\LARGE \bf Appendix}

\section{Qualitative Results}
\begin{figure*}[h]
    \centering
    \includegraphics[width=\linewidth]{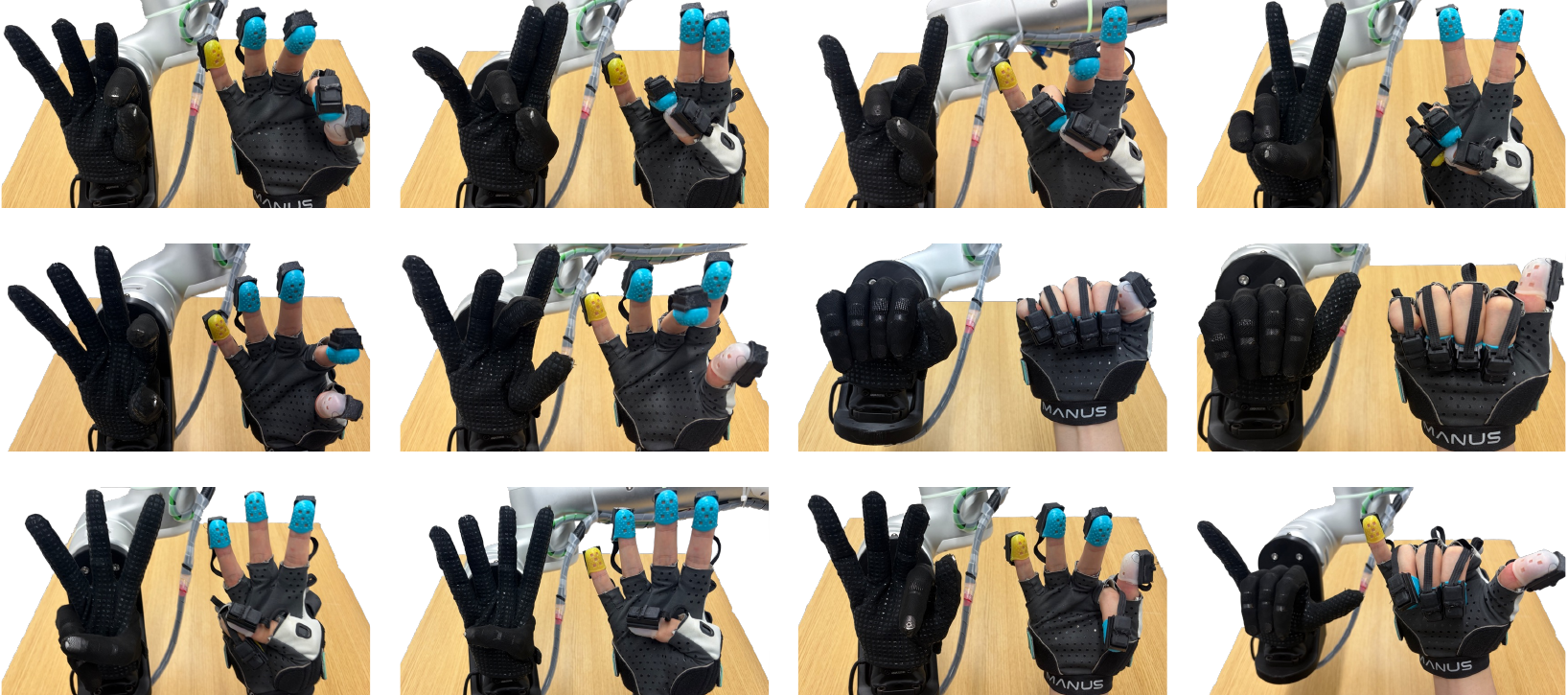}\vspace{-0.1cm}
    \caption{\textbf{Qualitative Results of \sys.} }
    \label{fig:qualitative}\vspace{-0.4cm}
\end{figure*}

Fig.~\ref{fig:qualitative} illustrates real-world qualitative results of \sys on Wuji Hand~\cite{wuji_hand}. We can observe that \sys provides precise and intuitive hand retargeting on multiple types of gestures.

\section{Implementation Details}
\paragraph{Networks} Both the finger mapper $f_m$ and the contact classifier $f_c$ are implemented using MLPs. For a hand with $F$ fingers, we employ $F$ sub-nets for $f_m$ and $(F-1)$ sub-nets for $f_c$. All sub-nets of $f_m$ have the same size of $(3,128,128,3)$, whose input and output are human fingertip positions and retargeted fingertip positions, respectively. All sub-nets of $f_c$ have the same size of $(6,128,128,1)$ followed by the Sigmoid function. The input of $f_c^i (i>0)$ is a fingertip position concatenation of the thumb and the $i$-th finger, and the output is a one-dimensional score distributed in $[0,1]$.

\paragraph{Data Preprocessing} For each human/robot finger, we centralize the fingertip position, compute the ranges along each axis, and select the axis with the maximum range. The selected range is then used to normalize the positions. The normalized positions are distributed in $[-1,1]$ without geometric distortion.

\paragraph{Training} The finger mapper $f_m$ and the contact classifier $f_c$ can be trained in parallel since they are independent from each other. Both networks are trained with the learning rate of 0.0001, and the batch size of 2048. During the training of $f_m$, we randomly perturb each sample with a generated delta movement and compute $\mathcal{L}_\text{motion}$ using the jittered positions. Anchor data with the batch size of 32 are also fed into $f_m$ for the computation of $\mathcal{L}_\text{align}$. The loss for training $f_m$ is $\mathcal{L}_\text{mapping} = \mathcal{L}_\text{P-Chamfer} + \mathcal{L}_\text{dist} + \mathcal{L}_\text{motion} + \mathcal{L}_\text{anchor}$, and we do not specifically tune the weights of each loss term.

\paragraph{Inference} The retargeted fingertip positions predicted by $f_m$ need to be transformed to joint configurations. We simply use nearest neighbor search (NNS) to get the corresponding joint configurations from the given fingertip positions and the last joint configurations. It can also be implemented using inverse kinematics or neural networks, but we find NNS is sufficient for our experiments. We use a threshold of $0.9$ for the scores predicted by $f_c$. For the predicted contact pattern of two fingers, we simply search the nearest fingertip positions from pre-generated templates using the corresponding positions predicted by $f_m$.

\section{Real-World Teleoperation}
\subsection{Task Description}
\paragraph{Spray-Bottle Triggering (\textit{Sprink})} Grasp the spray-bottle with thumb, middle, ring, and pinky fingers. Trigger the bottle with the index finger, then release it. This task involves both power grasping and in-hand manipulation.

\paragraph{Light-Bulb Screwing (\textit{Screw})} Use the thumb and index finger to screw in the light-bulb to turn it on. The progress is finished when the light is stable without flashing. This task is designed to evaluate the intuitiveness of the retargeting methods.

\paragraph{Steak Shoveling (\textit{Shovel})} Grasp the spatula with the full hand, use it to shovel up the steak, pour it into the bowl, and then place the spatula back. The task difficulty mainly depends on the stability of the power grasp.

\paragraph{Small-Ball Picking (\textit{Pick-10})} Use the thumb and index finger to grasp the small balls in the basket and place them into the bowl until all 10 balls are successfully transferred. This task is designed to evaluate the accuracy of the retargeted pinch poses. The pinch success rate is computed as $10/A$ where $A$ denotes the number of attempts.

\begin{wrapfigure}{r}{0.5\linewidth}\vspace{-0.4cm}
    \centering
    \includegraphics[width=\linewidth]{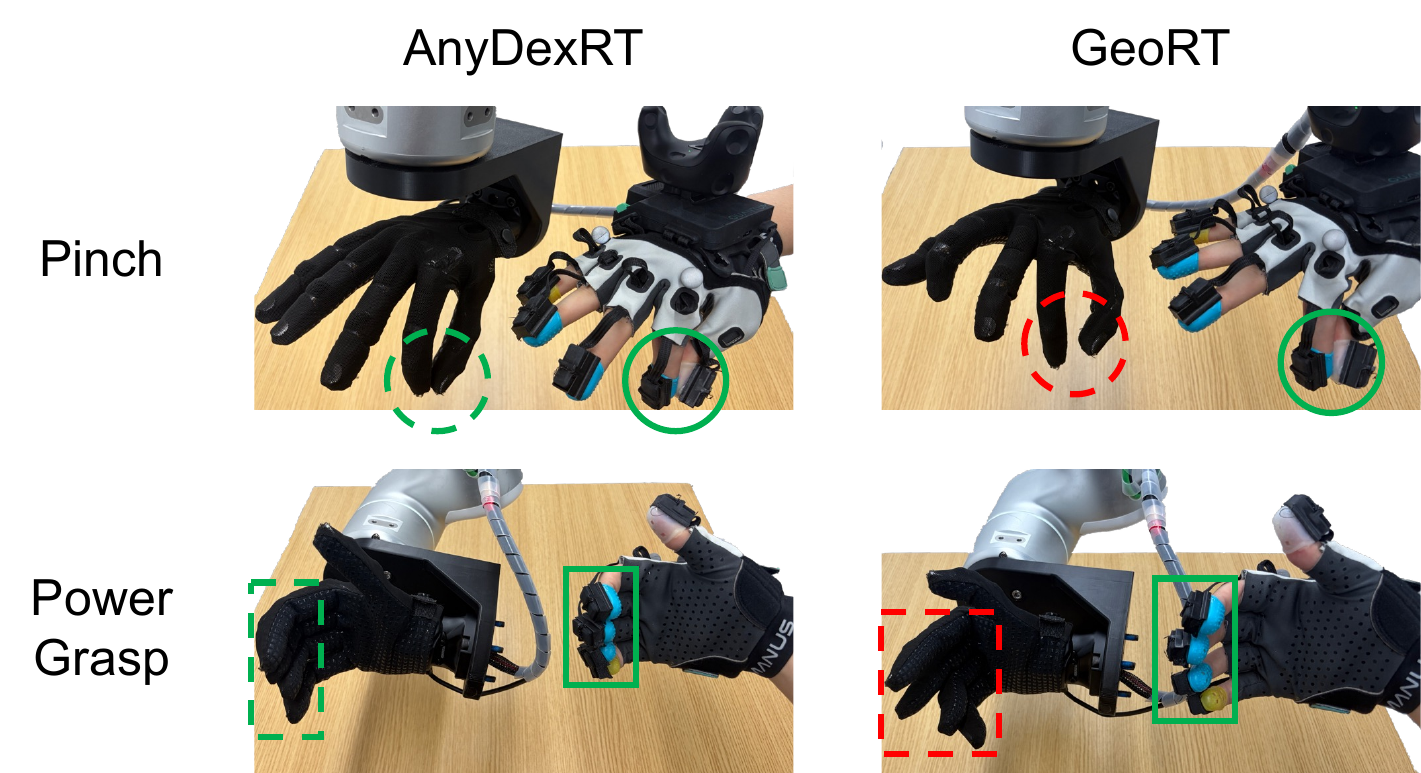}\vspace{-0.1cm}
    \caption{\textbf{Analysis of Real-World Teleoperation.}}
\label{fig:real_analysis}\vspace{-0.4cm}
\end{wrapfigure}
\subsection{Case Analysis}
We compare the retargeted pinch poses and power grasps output by \sys and GeoRT~\cite{geort} in Fig.~\ref{fig:real_analysis}. The accuracy of pinch poses decides the operation speed in \textbf{\textit{Pick-10}}, and the stability of power grasps ensures the success of \textbf{\textit{Sprink}} and \textbf{\textit{Shovel}}. GeoRT fails to predict accurate poses due to the widely distributed redundant fingertip space of the Wuji Hand. With human guidance, \sys successfully reduces this ambiguity and improves the intuitiveness of teleoperation.

\begin{wrapfigure}{r}{0.5\linewidth}\vspace{-0.4cm}
    \centering
    \includegraphics[width=\linewidth]{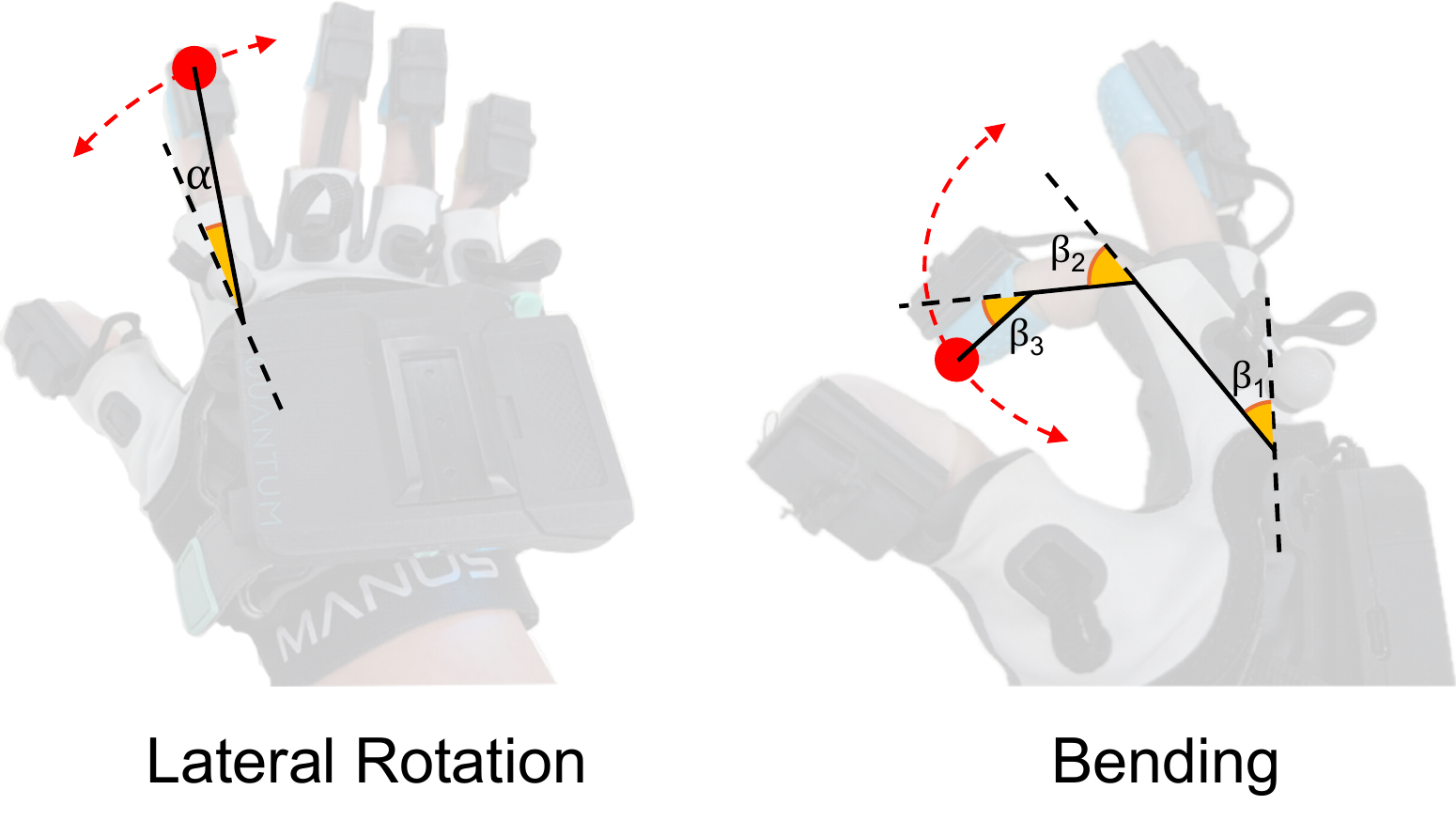}\vspace{-0.1cm}
    \caption{\textbf{Anchor Types and Collection.}}
\label{fig:anchor_type}\vspace{-0.4cm}
\end{wrapfigure}
\section{Anchor Data Collection}\label{appendix:anchor}
\subsection{Anchor Type}
The anchors can be defined according to the operators' habits. In Fig.~\ref{fig:anchor_type}, we provide the configuration in our experiments, which contains two types of paired anchors: \textbf{lateral rotation} and \textbf{bending}.

\paragraph{Lateral Rotation} Let $\beta_1 = \beta_2 = \beta_3 = 0$, we rotate the finger by changing $\alpha$. This type of data can be easily collected by putting one's hand on the table.

\paragraph{Bending} Let $\alpha=0$, the finger is rotated by changing $\beta_1$, $\beta_2$ and $\beta_3$. To simplify the collection process, we set $\beta_1 = \beta_2 = \lambda\beta_3$. According to~\cite{rijpkema1991computer}, $\lambda$ is usually a constant in bending, and we use $\lambda = 2$ in our experiments. Operators may use other constants (\cite{rijpkema1991computer} uses $3/2$) based on the characteristics of their own hands. It's hard to precisely define and collect the bending anchor for the thumb, so we just generate a rough bending trajectory of the robotic thumb finger and let operators imitate the finger pose. We find it is sufficient to conduct the task in our experiments.

In our experiments, we collect the bending anchors for all five fingers, and lateral rotation anchors for the thumb only. The same set of human anchors is shared across all robotic hands during training.

\subsection{Collection Process}
We describe the collection process for anchor data as follows:
\begin{enumerate}
    \item[\textbf{(P1)}] \textbf{Human Anchor Collection.}
    We collect each type of human anchor by repeatedly performing the corresponding finger motion and recording the fingertip trajectories. The motion ranges of both $\alpha$ and $\beta_1$ are $[0, \pi/2]$. For the four non-thumb fingers, bending trajectories are collected simultaneously through repeated fist-closing motions. The two types of thumb anchors are collected separately using the same procedure.
    \item[\textbf{(P2)}] \textbf{Robot Anchor Generation.}
    The robot anchors are generated in simulation by uniformly sampling $K$ joint configurations and recording the corresponding fingertip trajectories. The joint limits used for trajectory generation are the same as those specified in \textbf{(P1)}.
    \item[\textbf{(P3)}] \textbf{Human-Robot Anchor Pairing.}
    The collected human anchors are fitted with 2D curves, from which we uniformly sample $K$ fingertip positions to establish one-to-one correspondence between human and robot anchors.
    
\end{enumerate}

During the process, only \textbf{(P1)} needs human participation, while \textbf{(P2)} and \textbf{(P3)} are executed autonomously by computers. Note that human guidance is designed to make teleoperation more intuitive according to the operator's own habits. Therefore, the operator does not need to strictly align their finger with the defined anchor positions, and slight offsets are acceptable in practice.

\section{Update Notes}

\paragraph{Parameter Correction} In the arXiv v1 version, an incorrect parameter setting for robot anchor generation ($\lambda = 1$ instead of 2) led to an unexpected degradation in GMC performance. After correcting this issue, \sys consistently outperforms the baselines on both metrics. The definition of $\lambda$ is provided in Appendix~\ref{appendix:anchor}.

\paragraph{Anchor Collection Optimization} We simplify the human anchor collection process and automate human-robot anchor pairing. Further details are provided in Appendix~\ref{appendix:anchor}.

\end{document}